\documentclass{article} 
\usepackage{iclr2024_conference,times}

\usepackage{amsmath,amsfonts,bm}

\def\eqref#1{equation~\ref{#1}}

\def\1{\bm{1}}

\DeclareMathAlphabet{\mathsfit}{\encodingdefault}{\sfdefault}{m}{sl}
\SetMathAlphabet{\mathsfit}{bold}{\encodingdefault}{\sfdefault}{bx}{n}

\usepackage{hyperref}
\usepackage{url}
\usepackage{subcaption}
\usepackage{tikz}
\usepackage{algorithm2e}
\usepackage{wrapfig}
\usepackage{multirow}
\usepackage{multicol}
\usepackage{makecell}
\usepackage{tablefootnote}
\usepackage{dblfloatfix}

\title{Unsupervised Domain Adaptation within Deep Foundation Latent Spaces}

\author{Dmitry~Kangin, \& Plamen~Angelov \\
School of Computing and Communications\\
Lancaster University\\
 \\
\texttt{\{d.kangin1@,p.angelov\}@lancaster.ac.uk} \\
}

\iclrfinalcopy 
\begin{document}

\maketitle

\begin{abstract}

The vision transformer-based foundation models, such as ViT or Dino-V2, are aimed at solving problems with little or no finetuning of features. Using a setting of prototypical networks, we analyse to what extent such foundation models can solve unsupervised domain adaptation without finetuning over the source or target domain. Through quantitative analysis, as well as qualitative interpretations of decision making, we demonstrate that the suggested method can improve upon existing baselines, as well as showcase the limitations of such approach yet to be solved. 
\end{abstract}

\section{Introduction}

 \begin{figure}[!b]
\centering{
    \includegraphics[width=0.75\textwidth]{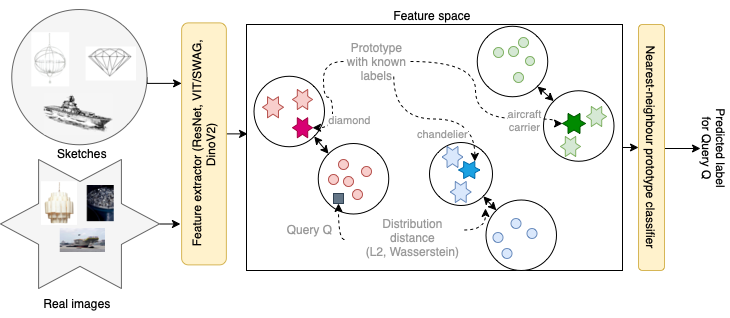}
}
\caption{The methodology scheme: (1) the images from multiple domains (e.g., sketches and real images) are embedded into the feature space and, for each domain, separately clustered using $k$-means. The cluster centroids for one of the domains ('source domain'), shown in bright colour in the figure and referred to as 'prototypes', are provided with labels. (2) Domain adaptation is performed through inter-domain cluster matches with $\ell^2$ or Wasserstein distance. (3) Decision making through nearest-neighbour prototype classifier performs the prediction}
\label{overall_scheme}
\end{figure}

With the advancement of foundation models, improvements in semi- and unsupervised learning methods can shift from end-to-end training towards decision making over the foundation models' latent spaces (\cite{oquab2023dinov2,angelov2023towards}). 

Below we describe the problem of unsupervised domain adaptation (UDA) (\cite{saenko2010adapting}). Consider a \textit{source} image dataset $\mathcal{S} = \{I^\mathcal{S}_{1}, \ldots I^\mathcal{S}_{n}\}$ and a \textit{target} dataset $\mathcal{T} = \{I^\mathcal{T}_{1}, \ldots I^\mathcal{T}_{m}\}$. These datasets share the same set of classes $\{C_1 \ldots C_k\}$, however the training labels are only known for the source dataset.  The problem is, given a classifier trained on a source dataset, to adapt it,  without any target data labels, to classify data on a target dataset.

Many of the existing works targeting UDA focus on representation learning approach towards assimilating, in the feature space, the source and target data (\cite{saenko2010adapting}). For many contemporary works, this is performed using adversarial training or minimising distribution divergence to match the distributions between the source and the target domain (\cite{peng2019moment}).

However, such training may not be an option if one wants to avoid finetuning of the latent feature spaces. We address it by using prototypical networks (\cite{snell2017prototypical,chen2019looks,angelov2020towards}), which recast decision making process into a function of prototypes (\cite{angelov2023towards}), derived from the training data.  For this purpose, we combine the prototypical network within the latent feature space with distribution matching between source and target domains. In Figure \ref{overall_scheme}, we summarise how one can address UDA problem using a simple combination of clustering and distribution matching  within the latent feature spaces using L2 or Wasserstein distances. We show that such approach can lead to promising results on a number of problems and outcompete purpose-finetuned models for UDA; furthermore, it allows to interpret the reasons behind misclassification using geometric proximity analysis through latent feature spaces. 

\section{Related work}

\paragraph {Unsupervised domain adaptation} \cite{saenko2010adapting} described a problem of adaptation to visual domain shifts, where the model is trained on the source domain  and tested on the target one, with changes in lighting, background, viewpoint, amongst others.  The current methods to solve this problem use a variety of approaches including adversarial assimilation of source and target distribution \cite{tzeng2017adversarial, zhao2018multiple, liu2021adversarial,saito2018maximum}, moment matching \cite{zellinger2017central,peng2019moment}. \cite{peng2019moment} proposes a dataset for UDA and the problem statement for multi-source domain adaptation, where they evaluate the transfer  learning when there are multiple source domains.

\paragraph {Visual Transformers} Building upon the applications of the attention models to the natural language processing (\cite{vaswani2017attention}), vision transformers (\cite{dosovitskiy2021an}) allowed not only to improve the performance but also provide generalisation capabilities (\cite{zhang2022delving}, \cite{oquab2023dinov2}). A substantial amount of literature is devoted to the aspects of pretraining (\cite{singh2022revisiting}), architecture design (\cite{liu2021swin}), unsupervised representation learning (\cite{oquab2023dinov2}), and task-specific finetuning of vision transformers (\cite{dai2021transmed}). 

\paragraph {Transparency of decision making} The pursuit of analysis of existing deep learning models has led to a number of methods targeting aspects of transparency such as \textit{ante hoc}, by-design, interpretability and \textit{post hoc} explainability. The former has been manifested by the interpretable-through-prototype methods such as ProtoPNet (\cite{chen2019looks}), xDNN (\cite{angelov2020towards}) and IDEAL (\cite{angelov2023towards}), as well as explainable architectures such as B-cos (\cite{bohle2022b}). The latter has widely used methods based on the sensitivity analysis for the input data, such as GradCAM (\cite{selvaraju2017grad}) and \cite{simonyan2014deep}. In this work, we focus on prototype-based methods, which allow to interpret the decision-making part through similarity of the query sample to the prototype, and therefore, our work is closely related to prototypical networks. 

\section{Methodology}

\begin{algorithm}
\caption{Proposed algorithm for UDA}\label{alg:one}
\KwData{ Source and target datasets $\mathcal{S}, \mathcal{T}$, Source query $q_s$, target query $q_t$ }
\KwResult{Source dataset class predictions $\hat{y}^{\mathcal{S}} (q_s)$, target dataset class predictions 
$\hat{y}^{\mathcal{T}} (q_t)$}
$\mathcal{S} \gets \phi(\mathcal{S}), \mathcal{T} \gets \phi(\mathcal{T})$ \tcp*[l]{$\phi$ denotes feature extractor}
$P_\mathcal{S}^c, C_{\mathcal{S}}\gets k$-means $(\mathcal{S}, | C_{\mathcal{S}}|)$ \tcp*[l]{Clustering of source data, $P_S^c$ are cluster centroids, $C_{\mathcal{S}}$ are cluster subsets of $\mathcal{S}$}
$P_\mathcal{S} \gets \{\arg \min_{s \in \mathcal{S}} \ell^2 (s, p) \forall p \in P_\mathcal{S}^c\}$\tcp*[l]{Selecting closest prototypes from $\mathcal{S}$}
$P_\mathcal{T}^c, C_{\mathcal{T}} \gets k$-means $(\mathcal{T},  |C_{\mathcal{T}}|)$ \tcp*[l]{Clustering of target data, $P_\mathcal{T}^c$ are cluster centroids, $C_{\mathcal{T}}^c$ are cluster subsets of $\mathcal{T}$}
$P_\mathcal{T} \gets \{\arg \min_{t \in \mathcal{T}} \ell^2 (t, p) \forall p \in P_\mathcal{T}^c\}$\tcp*[l]{Selecting closest prototypes from $\mathcal{T}$}
$L_\mathcal{S}\gets \mathtt{RetrieveLabels} (P_\mathcal{S})$ \tcp*[l]{Retrieve labels for every source prototype from $P_\mathcal{S}$ (and its corresponding cluster)}
$D_{\mathcal{ST}} = \mathtt{DistanceMatrix} (C_\mathcal{S}, P_\mathcal{S}, C_\mathcal{T}, P_\mathcal{T})$ \tcp*{Wasserstein or $l^2$ centroid distance matrix between clusters}
$L_T \gets \mathtt{ClosestMapping} (D_{ST}, L_S)$ \tcp*{Target clusters receive labels of the closest source cluster according to the distance matrix $D_{\mathcal{ST}}$}

$\hat{y}(q_s; P_\mathcal{S}) \gets L_\mathcal{S}[\arg \min_{p\in P_\mathcal{S}} \ell^2(p, q_s)]$  \tcp*{Source query class prediction}
$\hat{y}(q_t; P_\mathcal{T}) \gets L_T[\arg \min_{p\in P_\mathcal{T}} \ell^2(p, q_t)]$  \tcp*{Target query class prediction}
\end{algorithm}
In Algorithm \ref{alg:one}, we present the methodology for the proposed analysis. The methodology is split into (1) feature extraction, (2) selection of prototypes through clustering,  (3)  matching the prototypes between the source and the target domain (4) source and target class prediction.

We use two distinct methods for measuring distance between the clusters: $l^2$ distance between cluster prototypes, and the Sinkhorn approximation of the 2-Wasserstein distance (see details of implementation in the Appendix \ref{experimental_setup_appendix}).




\section{Experiments}

\paragraph{Experimental conditions} We reproduce the experimental setting from \cite{peng2019moment}, with the further experimental details described in Appendix \ref{experimental_setup_appendix}. While in \cite{peng2019moment} the model is proposed for the multi-source UDA, we present the results only for a single source domain.  The dataset contains six domains: sketch (ske), real (rel), quickdraw (qdr), painting (pnt), infograph (inf), and clipart(clp), each split into the same $345$ categories of common objects. 

\begin{figure}[]    
    \centering
    {
    \begin{subfigure}{\textwidth}
    \includegraphics[width=\textwidth]{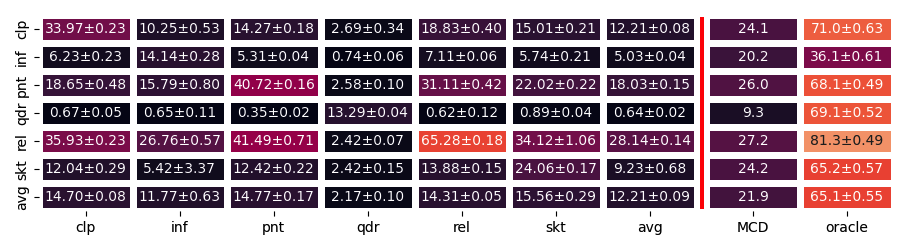}
    \caption{ResNet-152, finetuned on ImageNet, MCD (\cite{saito2018maximum}) and oracle results are taken from \cite{peng2019moment}}
    \label{tab:UDADomainNet_ResNet_152_ImageNet1K}       
    \end{subfigure}
    }
    \begin{subfigure}{0.75\textwidth}
    \centering
    {
    \includegraphics[width=\textwidth]{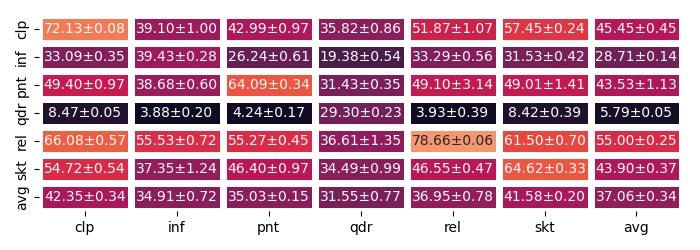}
    \caption{SWAG ViT-H/14 without finetuning ($l^2$ cluster prototypes distance)}
    \label{tab:UDADomainNet_ViT_H14}
    }
    \end{subfigure}
    \begin{subfigure}{0.75\textwidth}
    \centering
    \includegraphics[width=\textwidth]{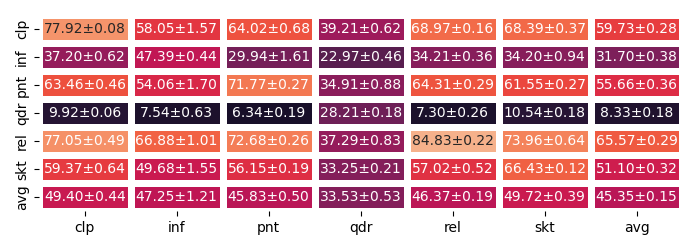}
    \caption{ SWAG ViT-H/14, ImageNet1K finetuning ($l^2$ cluster prototypes distance)}
    \label{tab:UDADomainNet_Vit_H14_ImageNet1K}
    \end{subfigure}
        \begin{subfigure}{0.75\textwidth}
    \centering
    {
    \includegraphics[width=\textwidth]{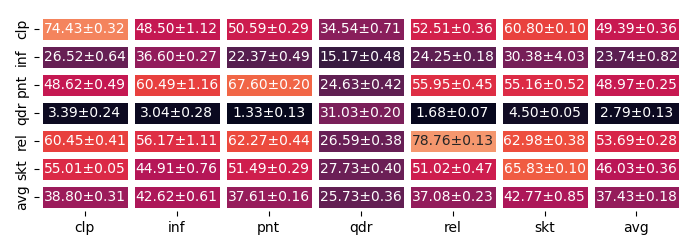}
    \caption{DinoV2 ViT-G/14, no finetuning ($l^2$ cluster prototypes distance)}
    \label{tab:UDADomainNet_DinoV2_ViT_G14_ImageNet1K}  
    }
    \end{subfigure}
    \begin{subfigure}{0.75\textwidth}
    \centering
    \includegraphics[width=\textwidth]{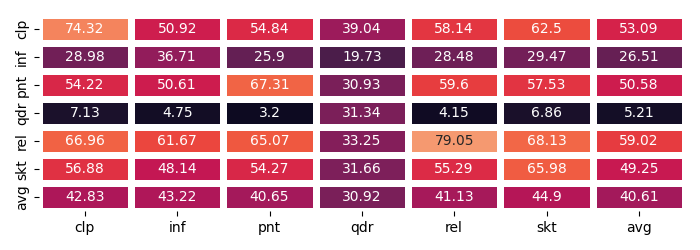}
    \caption{ DinoV2 ViT-G/14 (Sinkhorn approximation of the Wasserstein distance)}
    \label{tab:UDADomainNet_DinoV2_G14_Sinkhorn}
    \end{subfigure}
    \caption{UDA results for  different backbone architectures (columns denote source domains, and rows denote target domains), $k$-means clustering ($5 \times 345=1725$ clusters, $345$ classes)}
\end{figure}

Table \ref{tab:UDADomainNet_ResNet_152_ImageNet1K} shows the performance of the proposed methodology for ResNet-152-based model, as well as for the oracle (purpose-fit model), based on ResNet-152, and for the MCD (\cite{saito2018maximum}) baseline, based on ResNet-101 (\cite{he2016deep}), which has shown the best performance in the analysis of \cite{peng2019moment}. It can be seen that in such scenario the model performs substantially worse than the state-of-the-art UDA method MCD (\cite{saito2018maximum}). In Tables \ref{tab:UDADomainNet_ViT_H14}-\ref{tab:UDADomainNet_DinoV2_ViT_G14_ImageNet1K}, we demonstrate that the performance improves for various ViT backbones and ultimately overtakes MCD baseline (\cite{saito2018maximum}). It happens, however, that superior performance in transfer learning for Dino-V2 (\cite{oquab2023dinov2}) does not translate into a better performance on this UDA task. Furthermore, the version, pretrained on ImageNet-1K (Table \ref{tab:UDADomainNet_Vit_H14_ImageNet1K}) surprisingly shows better performance on practically every single task comparing to the counterpart without pretraining (Table \ref{tab:UDADomainNet_ViT_H14}). Comparison with the Wasserstein distance version of the method demonstrates that while the choice of distance shows some potential to improve the performance, it still lags behind ViT H/14, finetuned on ImageNet-1K.

In Section \ref{interpretability_analysis}, we demonstrate the analysis of the errors in the latent space by visualising the nearest prototypes for SWAG ViT-H/14, finetuned on ImageNet-1K. Such analysis demonstrates remarkable cross-domain generalisation, with a number of justified semantically meaningful errors (for example, mistaking diamonds for similarly-looking blueberries), as well as highlights that some of the errors are caused by preferring texture to the semantics (e.g., mistaking octagon for hexagon).

\section{Conclusion}
We show that, with just fixed ViT feature representation and domain distribution matching, one can solve a number of unsupervised domain adaptation problems within the foundational feature space, exceeding the performance of purpose-built representation learning frameworks for UDA, such as MCD (\cite{saito2018maximum}) and thus becoming a plausible alternative to finetuning-based domain adaptation.  In many cases, however, even though the results are wrong from the benchmark's point of view, they still can constitute a plausible answer even for the humans (see Figure \ref{interpretability_domain} in the Appendix). The described methodology can serve as a benchmark for evaluating such representation learning.  While foundation models provide competitive performance against purpose-built models in this challenge, some of the best-performing models, such as DinoV2 G/14, do not improve upon the performance of older models with lower general-purpose performance, such as SWAG-ViT, finetuned on ImageNet-1K. 

\section*{Acknowledgement}
This work is supported by ELSA – European Lighthouse on Secure and Safe AI funded by the European Union under grant agreement No. 101070617. Views and opinions expressed are however those of the author(s) only and do not necessarily reflect those of the European Union or European Commission.Neither the European Union nor the European Commission can be held responsible.

The computational experiments have been powered by a High-End Computing (HEC) facility of Lancaster University, delivering high-performance and high-throughput computing for research within and across departments.

\bibliography{iclr2024_conference}
\bibliographystyle{iclr2024_conference}

\appendix
\section{Experimental setup}
\label{experimental_setup_appendix}
We use the pretrained SWAG-ViT (\cite{singh2022revisiting}) models from publicly available repository\footnote{https://github.com/facebookresearch/SWAG}.  SWAG ViT H/14 model corresponds to the pretrained model with no finetuning. For the experiments we use NVIDIA RTX A2000 12GB powered workstation. 
sklearn implementation of $k$-means clustering is used.

All the experiments are repeated with three different random seeds to calculate the confidence interval, except from the Wasserstein distance experiment in Figure \ref{tab:UDADomainNet_DinoV2_G14_Sinkhorn} which do not include confidence interval computation due to the computational costs.

For calculating the Sinkhorn approximation of 2-Wasserstein distance, we use the following \texttt{geomloss} library function: \texttt{geomloss.SamplesLoss(loss='sinkhorn', p=2, blur=1e-5)}. 

\section{Interpretability analysis}
\label{interpretability_analysis}
 \begin{figure}
  \centering{

\begin{subfigure}{\textwidth}
\includegraphics[width=0.45\textwidth]{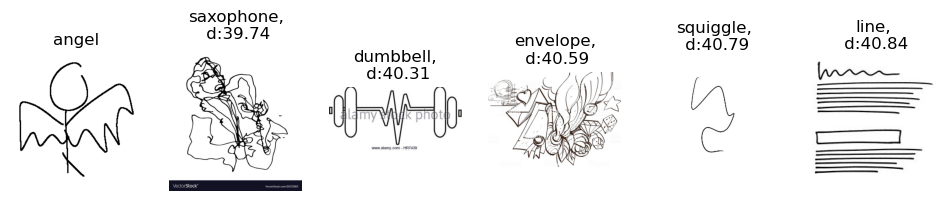}\hfill\vline\hfill
\includegraphics[width=0.45\textwidth]{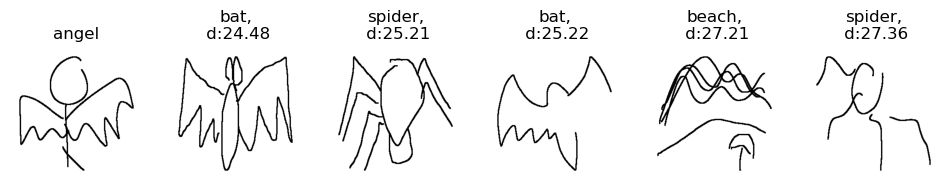}
\caption {Angel, recognised incorrectly (testing, sketch $\rightarrow$ quickdraw), left: \textit{sketch}, right: \textit{quickdraw}}
\label{domains:prototypes_angel}
\end{subfigure}

\begin{subfigure}{\textwidth}
\includegraphics[width=0.37\textwidth]{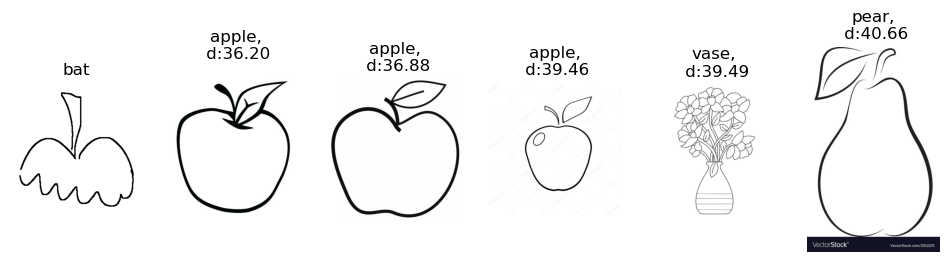}
\hfill\vline\hfill \includegraphics[width=0.49\textwidth]{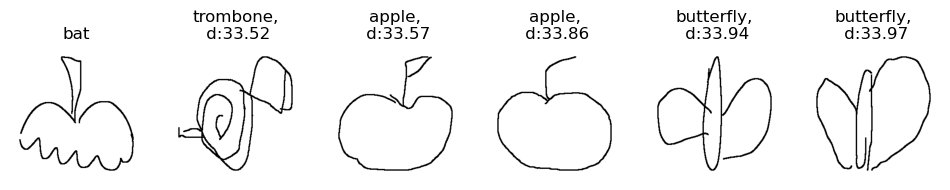}
\caption {Bat, recognised incorrectly (testing, sketch $\rightarrow$ quickdraw), left: \textit{sketch}, right: \textit{quickdraw}}
\label{domains:prototypes_bat}
\end{subfigure}

\begin{subfigure}{\textwidth}
\includegraphics[width=0.43\textwidth]{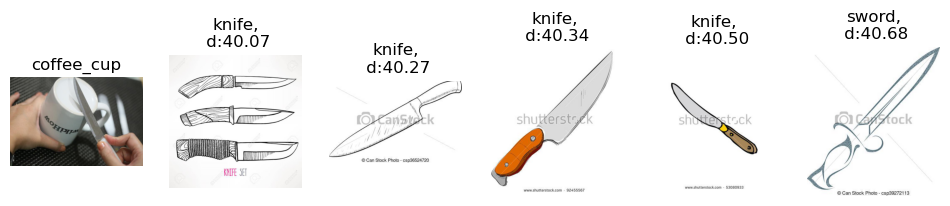} \hfill\vline\hfill 
\includegraphics[width=0.49\textwidth]{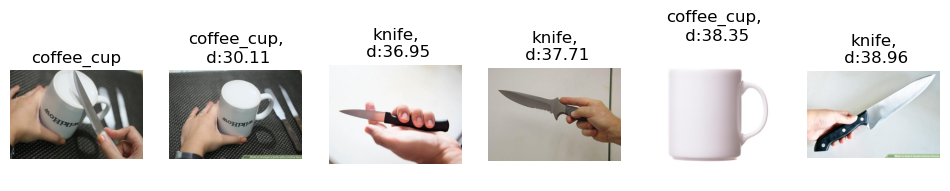}
\caption {Coffee cup (testing, sketch $\rightarrow$ real), left: \textit{sketch}, right: \textit{real}}
\label{domains:prototypes_coffee_cup}
\end{subfigure}

\begin{subfigure}{\textwidth}
\includegraphics[width=0.4\textwidth]{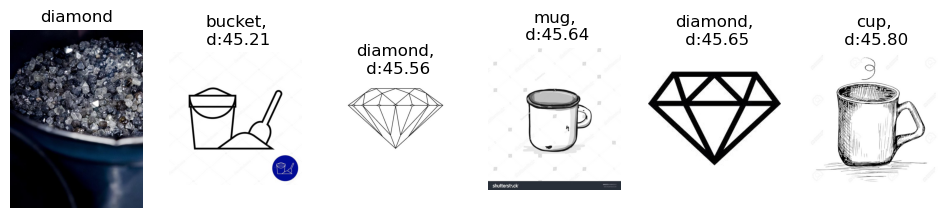}\hfill\vline\hfill
\includegraphics[width=0.4\textwidth]{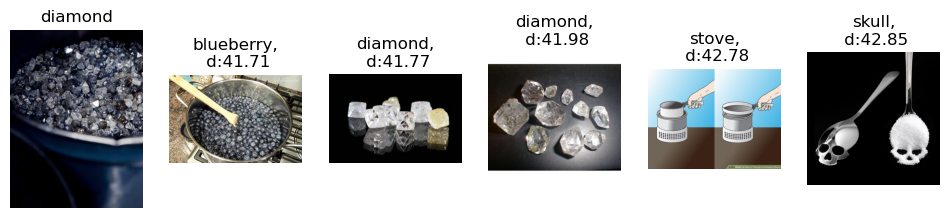}
\caption {Diamond, recognised incorrectly (testing, sketch $\rightarrow$ real), left: \textit{sketch}, right: \textit{real}}

\label{domains:prototypes_diamond}
\end{subfigure}

\begin{subfigure}{\textwidth}
\includegraphics[width=0.4\textwidth]{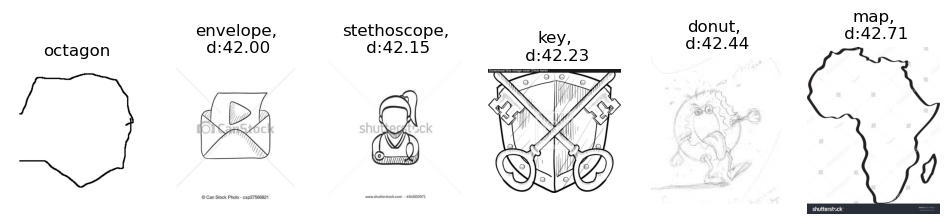}\hfill\vline\hfill
\includegraphics[width=0.49\textwidth]{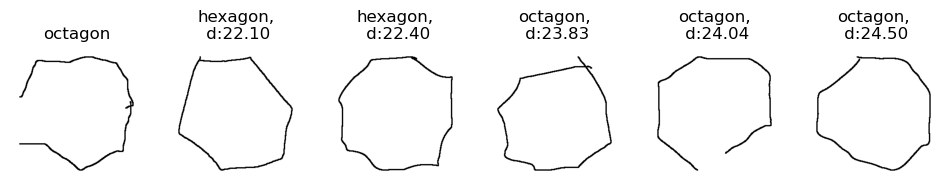}
\caption {Octagon, recognised incorrectly (testing, sketch $\rightarrow$ quickdraw), left: \textit{sketch}, right: \textit{quickdraw}}
\label{domains:prototypes_octagon}
\end{subfigure}

  \centering{
  \begin{subfigure}{\textwidth}
\includegraphics[width=0.45\textwidth]{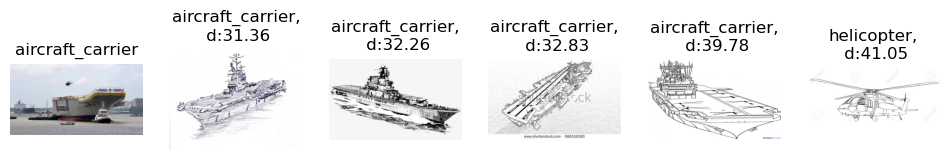} \hfill\vline\hfill
\includegraphics[width=0.45\textwidth]{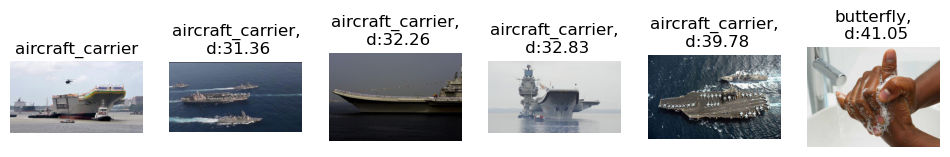}

\caption {Aircraft carrier (testing, sketch $\rightarrow$ real), left: \textit{sketch}, right: \textit{real} prototypes}
\label{domains:prototypes_aircraft_carrier}
\end{subfigure}
\begin{subfigure}{\textwidth}
\includegraphics[width=0.45\textwidth]{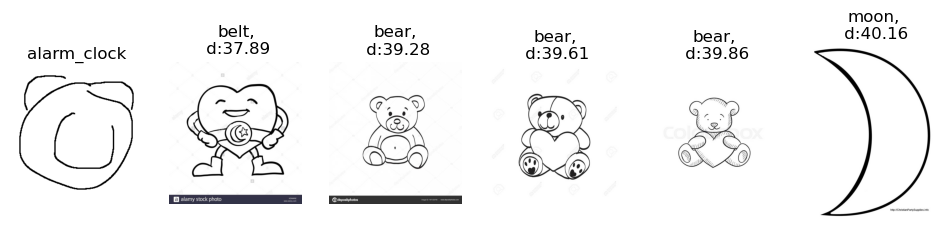} \hfill\vline\hfill
\includegraphics[width=0.49\textwidth]{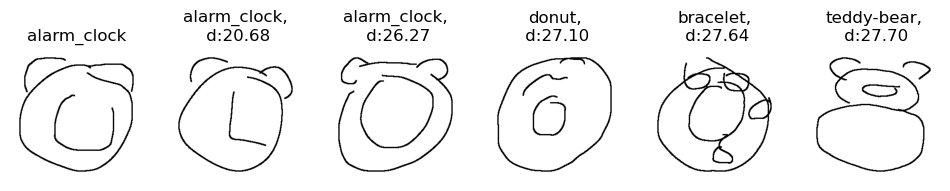}
\caption {Alarm clock (testing, sketch $\rightarrow$ quickdraw), left: \textit{sketch} prototypes, right: \textit{quickdraw} prototypes}
\label{domains:prototypes_alarm_clock}
\end{subfigure}
  
\begin{subfigure}{\textwidth}
\includegraphics[width=0.45\textwidth]{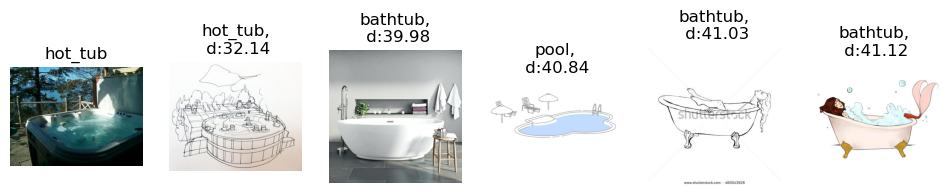}\hfill\vline\hfill
\includegraphics[width=0.45\textwidth]{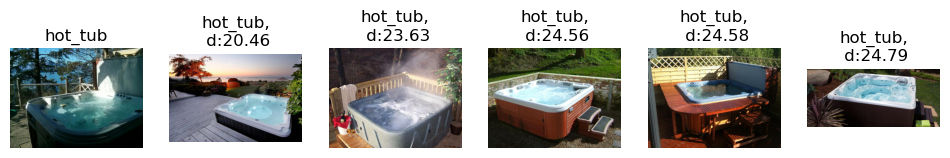}
\caption {Hot tub (testing, sketch $\rightarrow$ real), left: \textit{sketch} prototypes, right: \textit{real} prototypes}
\label{domains:prototypes_hottub}
\end{subfigure}

\begin{subfigure}{\textwidth}
\includegraphics[width=0.49\textwidth]{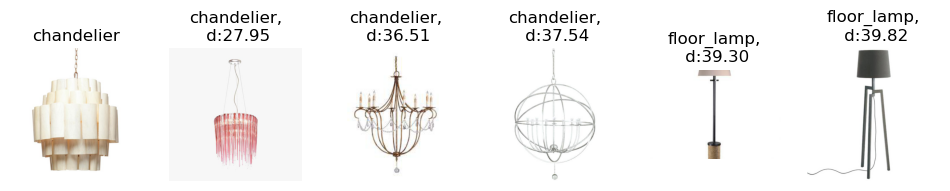}\hfill\vline\hfill
\includegraphics[width=0.45\textwidth]{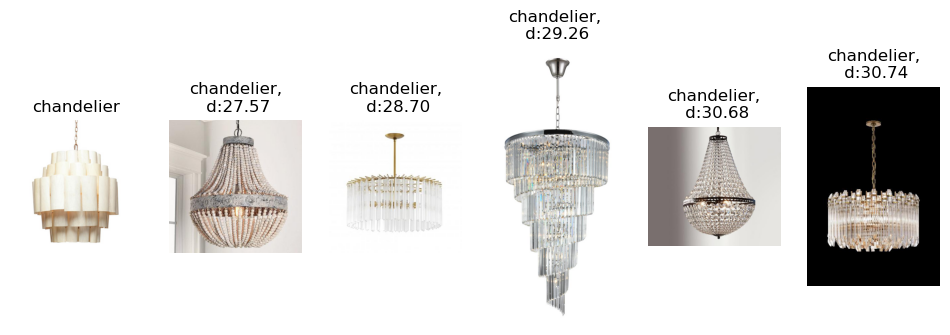}
\caption {Chandelier (training, sketch $\rightarrow$ real), left: \textit{sketch} prototypes, right: \textit{real} prototypes}
\label{domains:prototypes_chandelier}
\end{subfigure}

\begin{subfigure}{\textwidth}
\includegraphics[width=0.49\textwidth]{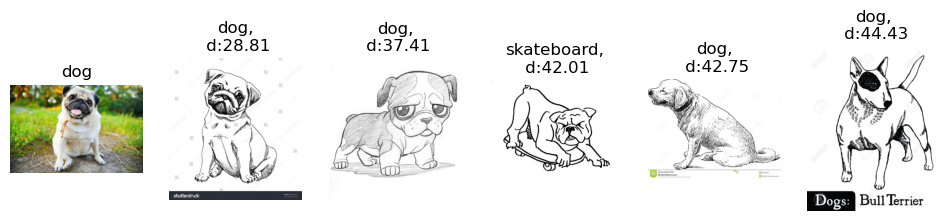}\hfill\vline\hfill
\includegraphics[width=0.49\textwidth]{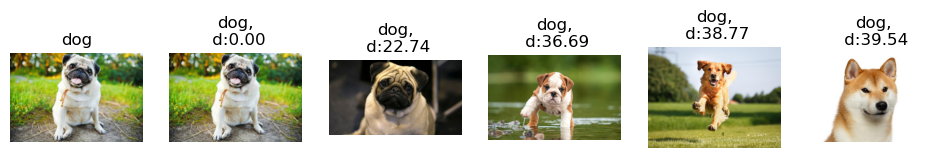}
\caption {Dog (training, sketch $\rightarrow$ real), left: \textit{sketch} prototypes, right: \textit{real} prototypes}
\label{domains:prototypes_dogs}
\end{subfigure}
}

}
\caption{Interpretations of decision making through closest prototypes (the leftmost images are queries, and the further ones are prototypes)}
\label{interpretability_domain}
\end{figure}

 To better understand the performance scores, we provide visualisation for nearest prototypes for the ViT-SWAG, finetuned on ImageNet-1K ((see Figure \ref{interpretability_domain}). For the same latent space, this numerical estimate can be compared like-for-like between different examples and datasets. 

One can see from Figure \ref{interpretability_domain}, that in many cases, even for the incorrect classification, the closest examples make semantic sense. In Figure \ref{domains:prototypes_aircraft_carrier}, the model successfully relates aircraft carriers in different poses, sketched or photographed.  The alarm clock (Figure \ref{domains:prototypes_alarm_clock}) recognises the quickdraw prototypes, but struggles to generalise to the sketch prototypes. Angel quickdraw prototypes (see Figure \ref{domains:prototypes_angel} ) show confusion (in the quickdraw domain) between similarly-looking angels, spiders and bats. The results are worse for cross-domain examples.  The quickdraw sample in Figure \ref{domains:prototypes_bat}, supposed to be a bat, looks indeed much like an apple. This is duly reflected by the model outputs. 
Coffee cup example (see Figure \ref{domains:prototypes_coffee_cup}), demonstrates competition between the coffee cup and the knife prototypes as the image contains both. Seeing the nearest blueberry example to the real diamond query image (see Figure \ref{domains:prototypes_diamond}) helps make sense out of the incorrect recognition. In the hot tub example (Figure \ref{domains:prototypes_hottub}), one can see that the varieties of appearances of the same objects are shown to have a low distance. The feature space also places closer different types of light fixtures, as one can see in Figure \ref{domains:prototypes_chandelier}, in sketch and real domains alike.The limitations of differentiating between geometric figures such as hexagons and octagons (see Figure \ref{domains:prototypes_octagon} suggest that despite generality, textural information often trumps the semantic one. For the dogs scenario on training data (see Figure \ref{domains:prototypes_dogs}, one can see that while the model has sometimes advantage to find the exact match with the prototypes, it also shows remarkable cross-domain generalisation.

\end{document}